\documentclass[10pt,twocolumn,letterpaper]{article}

\usepackage[pagenumbers]{cvpr} % To force page numbers, e.g. for an arXiv version

\newcommand{\ignore}[1]{}

\ifnum\pdfstrcmp{\jobname}{output}=0
\else
  \usepackage[english]{babel}
\fi

\usepackage{xcolor}
\usepackage{fancybox}
\usepackage{calc}
\usepackage{adjustbox}
\usepackage{pgfplots}
\usepackage{tikz}
\usetikzlibrary{patterns}
\pgfplotsset{compat=1.18}
\usepackage{xfp}

\makeatletter

\def\@divA{.0392\height}  % 87/2218
\def\@divB{.2769\height}  % 614/2218
\def\@divC{.5118\height}  % 1135/2218
\def\@divD{.7467\height}  % 1656/2218
\def\@divE{1.0\height}    % 2218/2218

\newcommand{\@stackimg}[4]{%
    \adjustbox{trim={0 0 {#1} 0}, clip, width=\linewidth}{%
        \begin{tabular}{@{}c@{}}%
            \adjustbox{trim={0 {\dimexpr#2\relax} 0 0}, clip}{\includegraphics{#4}}\\[-4pt]%
            \adjustbox{trim={0 0 0 {#3}}, clip}{\includegraphics{#4}}%
        \end{tabular}%
    }%
}

\newcommand{\fullcompfig}[3][none]{%
    \def\hcrop{0\width}%
    \ifnum#2=2\def\hcrop{.6009\width}\fi%
    \ifnum#2=3\def\hcrop{.4\width}\fi%
    \ifnum#2=4\def\hcrop{.1989\width}\fi%
    \def\nonetext{none}%
    \def\delframe{#1}%
    \ifx\delframe\nonetext%
        \ifnum#2=5\relax%
            \includegraphics[width=\linewidth]{#3}%
        \else%
            \adjustbox{trim={0 0 {\hcrop} 0}, clip, width=\linewidth}{\includegraphics{#3}}%
        \fi%
    \else%
        \ifnum#1=0\relax% Delete frame 0 (87-614): keep 0-87, then 614-2218
            \@stackimg{\hcrop}{\@divE-\@divA}{\@divB}{#3}%
        \else\ifnum#1=1\relax% Delete frame 1 (614-1135): keep 0-614, then 1135-2218
            \@stackimg{\hcrop}{\@divE-\@divB}{\@divC}{#3}%
        \else\ifnum#1=2\relax% Delete frame 2 (1135-1656): keep 0-1135, then 1656-2218
            \@stackimg{\hcrop}{\@divE-\@divC}{\@divD}{#3}%
        \fi\fi\fi%
    \fi%
}

\makeatother

\definecolor{cvprblue}{rgb}{0.21,0.49,0.74}
\usepackage[pagebackref,breaklinks,colorlinks,allcolors=cvprblue]{hyperref}
\usepackage{comment}

\def\paperID{13930} % *** Enter the Paper ID here
\def\confName{CVPR}
\def\confYear{2025}

\title{MotionV2V: Editing Motion in a Video}

\author{
  Ryan Burgert$^{1,2}$ ~~~~~~~ Charles Herrmann$^{1}$ ~~~~~~~ Forrester Cole$^{1}$ ~~~~~~~ Michael S Ryoo$^{2}$ \\  Neal Wadhwa$^{1}$ ~~~~~~~ Andrey Voynov$^{1}$ ~~~~~~~ Nataniel Ruiz$^{1}$\\
  \\
  $^{1}$Google\;\;\;\;   $^{2}$Stony Brook University
}

\begin{document}
\maketitle

\ignore{
\twocolumn[{
\renewcommand\twocolumn[1][]{#1}
\maketitle
\begin{center}
    \centering
    \vspace*{-.6cm}
    \vspace*{-.3cm}
    \captionof{figure}{Our method can edit videos in a true sense, where content is preserved but motion is changed. Above we show some practical applications.}
\label{fig:teaser}
\end{center}
}]

\begin{figure}[!h]
    \centering
    \caption{Conceptual overview of our motion editing approach. Users provide an input video along with source motion tracks (colored dots connected by lines, extracted from the input) and target motion tracks (user-specified desired motion). Lines indicate point trajectories while dot presence/absence indicates visibility. Our diffusion model generates an output video matching the target motion. The method supports iterative editing: outputs can become inputs for subsequent edits, enabling complex sequential motion changes.}
    \label{fig:concept_overview}
    \vspace{-10pt}
\end{figure}
}

\begin{abstract}
While generative video models have achieved remarkable fidelity and consistency, applying these capabilities to video editing remains a complex challenge. Recent research has extensively explored motion controllability as a means to enhance text-to-video generation or image animation; however, we identify precise motion control as a promising, yet under-explored, paradigm for editing \textbf{existing videos}. In this work, we propose modifying video motion by directly editing sparse trajectories extracted from the input. We term the deviation between input and output trajectories a `motion edit' and demonstrate that this representation, when coupled with a generative backbone, enables many powerful video editing capabilities. To achieve this, we introduce a novel pipeline for generating `motion counterfactuals' — video pairs that share identical content but distinct motion — and fine-tune a motion-conditioned video diffusion architecture on this dataset. Our approach allows for edits that start at any timestamp and propagate naturally. In a 4-way head-to-head user study, our model achieves over 65\% preference against prior work. Please see our project page: \href{https://ryanndagreat.github.io/MotionV2V}{ryanndagreat.github.io/MotionV2V}
\end{abstract}

\section{Introduction}

\begin{figure*}
    \vspace*{-.6cm}
    \includegraphics[width=1.0\textwidth]{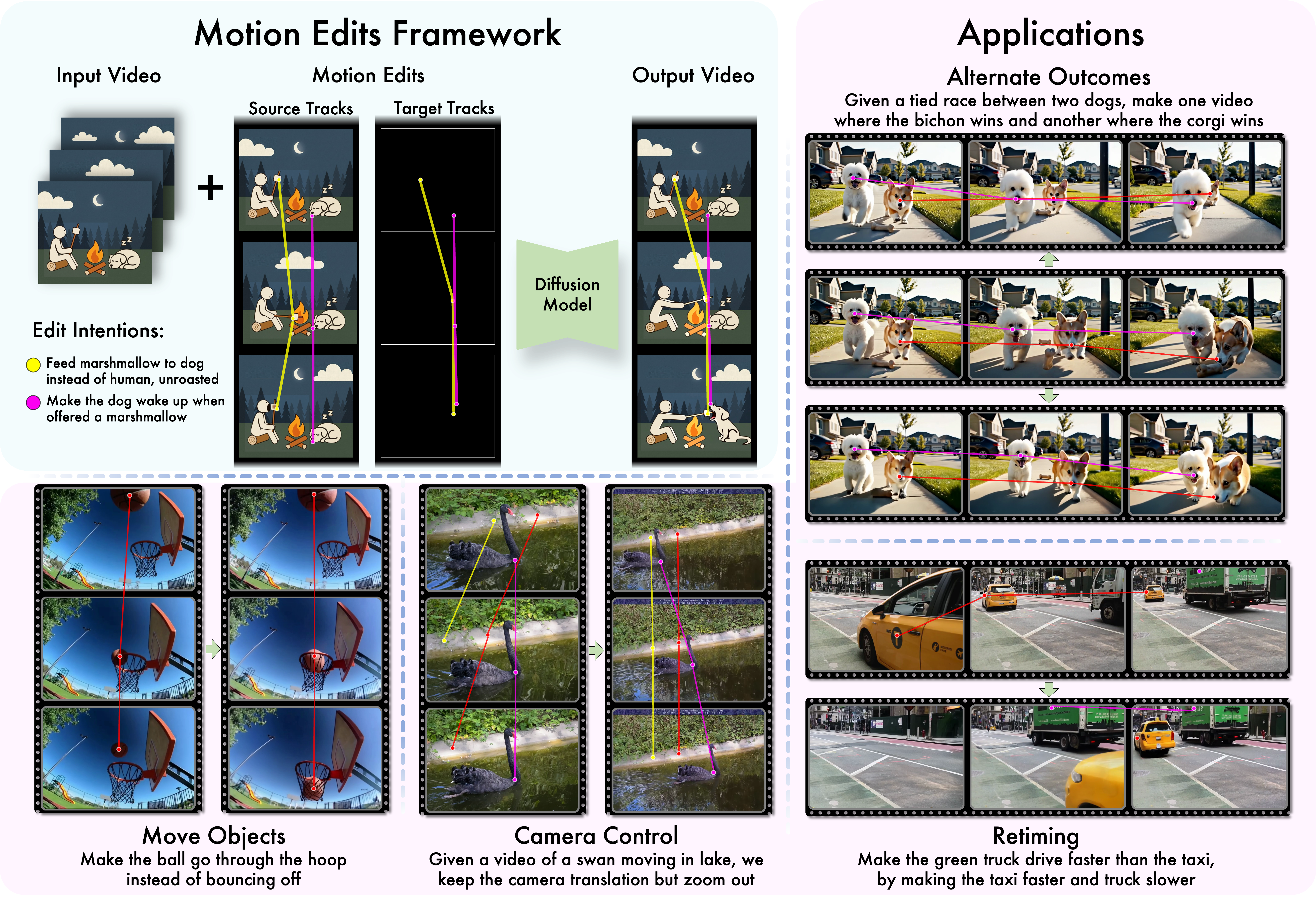}
    \vspace*{-0.8cm}
    \captionof{figure}{\textbf{Motion Edits Framework:} Users provide an input video along with source motion tracks (colored dots connected by lines, extracted from the input) and target motion tracks (user-specified desired motion). Lines indicate point trajectories while dot presence/absence indicates visibility. Our diffusion model generates an output video matching the target motion. \textbf{Applications:} Our method can edit videos in a true sense, where content is preserved but motion is changed.}
\label{fig:teaser}
\end{figure*}

Consider filming a climactic race between two dogs: your Corgi and a friend's Bichon. The original video sees the Bichon take the win. After countless recent advances in generative models, does the technology exist to modify this video such that your Corgi is victorious? We propose a method for generalized motion editing in existing user-provided videos that successfully tackles this unsolved problem.

Historically, tackling this problem in the VFX industry has been notably hard. A reshoot for scenes that need substantial changes is usually the necessary option. VFX pipelines can use tricks like retiming and plate stitching, isolated retimes with rotoscoping, or even full-dog CGI replacements. These typically require a high level of skill and large amounts of human hours.

Modern generative models, with their impressive priors, show promise in tackling traditional VFX tasks. In this subfield, current methods for motion editing fall into different categories, each exhibiting significant constraints. Image-to-video (I2V) based approaches like Re-Video~\cite{revideo2024} and Go-with-the-Flow~\cite{gowiththeflow2025} can only generate new video with specified motion conditioned on a single image. Using these on the first frame of a video can give the illusion of video motion control, but have significant drawbacks. For example, content generated in regions that do not appear in that initial frame will be entirely hallucinated, whereas for true video motion editing these regions are known and should remain identical. Re-Video attempts to address this problem by inpainting information from the original video into the edited video, a technique which fundamentally breaks down when the video includes camera movement. Human-specific methods like MotionFollower and MotionEditor can edit motion but are limited to full-body human movements and cannot handle general objects or scenes. Likewise there are also works that allow editing camera trajectories in videos such as ReCapture~\cite{recapture2024} and ReCamMaster~\cite{recammaster2025}. These are not able to edit subject motion.

In this work we introduce motion edits, a new approach for editing videos by controlling the change in motion from the original to the edited video using video diffusion models. While there has been a large amount of successful recent work on appearance-based video editing (i.e.\ transforming visual style while preserving motion structure), motion editing presents a fundamentally different challenge. When editing how objects move within a scene (e.g.\ making a person walk in a different direction), the structural correspondence between input and output videos is broken. This makes the problem harder than appearance-based video editing and renders standard video editing techniques like DDIM inversion ineffective.

% pause %
% Our method addresses this problem, and the limitations of prior work, by acting on the complete input video and a motion representation for the existing input video. Users provide an input video along with a small number of sparse tracking points placed on objects they want to control the motion of. These points are automatically tracked throughout the video and the user can then specify either to either keep them the same (in order to make sure they are in the same location in the edited video) or to change them (in order to change the motion). For example, consider a video of a person walking into a crowd. Our system would automatically place tracking points on both the person and crowd, allowing the user to, by preserving the tracks on the crowd and changing those on the person, specify a new direction for the person to walk.

Our method addresses this problem, and the limitations of prior work, by acting on the complete video and its motion representation. Users provide an input video along with some sparse tracking points on objects they wish to control; these objects are then automatically tracked throughout the video. Users can then choose to either anchor these points (to preserve original motion) or modify them (to edit the trajectory). For example, in a video of a person walking into a crowd, the system tracks both entities; the user can specify a new direction for the person by altering their tracks while strictly preserving the crowd’s original motion. In a more complex edit, the user can change the camera by editing all the points.

%Our novelties lie in two areas (1) we propose a method to generate synthetic counterfactual videos that share visual appearance but exhibit different motions, essentially generating high quality ``parallel realities'' where different motions occur under same initial (or intermediate) conditions and (2) a diffusion model architecture with carefully constructed motion conditioning trained on this data. After training, our model accepts videos and motion tracks with sparse points as input and yields a video with same appearance as the input video, but with the user-specified motion.

Our approach enables diverse video editing capabilities (Fig.~\ref{fig:teaser}): object motion editing, camera motion editing, control over the timing of content, and successive edits. The motion edits are naturally introduced into the output video and the video model handles plausibility correctly - e.g.\ when dragging a person's tracking point through an image the model will make the person traverse the image by walking. Our approach requires no manual masking and can handle any type of object, while also maintaining scene consistency even when both object motion and camera movement are edited simultaneously. And in contrast to prior methods, our approach also allows the ability to change when an object appears in the frame. Finally, our model generalizes to vastly different scenes and objects and achieves state-of-the-art performance in quantitative comparisons and human evaluations.
% Keeping this TODO, but we should do it succinctly and add it in the paragraph above. One sentence if possible.
% \TODO{Add detailed explanation of the editing workflow: how users initialize tracking points, how they edit trajectories in the GUI, and include a figure showing the editor interface.}

In summary, we propose the following contributions:
\begin{itemize}
\item We identify motion as a powerful control signal for video editing and propose directly editing sparse trajectories extracted from the input video to change the motion of the output video. We define the change between the input and target trajectories as a ``motion edit'' and show that motion edits, coupled with a powerful generative video model, can address several challenging video editing tasks. 
\item We present a methodology to train a video diffusion model to generate high quality ``motion counterfactual'' video pairs which have the scene appearance but different motion. As part of this, we also identify sources of data that work well in this training.
\item We propose a new model architecture with careful conditioning on both user-specified video and motion trajectories that generates a motion-edited output.
\end{itemize}

\label{sec:intro}

\label{sec:method}

----------------------------------------------------------

\section{Related Works}
% Diffusion models have revolutionized media generation, starting with foundational works on denoising diffusion probabilistic models in images generation~\cite{ddpm2020, stablediffusion2022} and rapidly extending to video~\cite{vdm2022, imagenvideo2022, makeavideo2022, bar2024lumiere}, and various other domains. Recent text-conditioned video models~\cite{cogvideox2024, wan, opensora2024} utilize transformers~\cite{dit2023} as the denoising architecture.

Diffusion models have fundamentally reshaped media generation, evolving from foundational image synthesis frameworks~\cite{ddpm2020, stablediffusion2022} to complex video dynamics~\cite{vdm2022, imagenvideo2022, makeavideo2022, bar2024lumiere}. Recent text-conditioned video models~\cite{cogvideox2024, wan, opensora2024} have further advanced the field by adopting transformer-based architectures~\cite{dit2023} for scalable denoising.

\subsection{Conditional Video Generation}
% Conditional video diffusion models extend the base text-to-video architectures with additional control signals beyond text and image. Inspired by the ControlNet~\cite{controlnet2023} architectural pattern, works have adapted the approach to video domains with various conditioning mechanisms~\cite{das2025, videocontrolnet2023, motioni2v2024} enabling video conditioning through depth maps, motion vectors, camera parameters, and other modalities.
% Other methods introduce V2V editing mechanisms by propagating edits across frames and preserving some features of an original video when generating an edited one~\cite{tokenflow2024, fatezero2023, codef2024, pix2video2023, tuneavideo2023, text2videozero2023, cove2024}. A series of works use the DDIM inversion-based approaches for appearance modifications~\cite{i2vedit2024, magicedit2023, stablevideo2023}. However, these methods are fundamentally designed for local appearance changes and cannot handle nonlocal motion edits where the structural correspondence between frames is broken. When motion patterns change, the temporal alignment assumptions underlying these approaches are violated.

Conditional video diffusion extends base text-to-video architectures by incorporating auxiliary control signals. Inspired by the spatial conditioning of ControlNet~\cite{controlnet2023}, recent works have adapted similar mechanisms to the temporal domain~\cite{das2025, videocontrolnet2023, motioni2v2024}, enabling guidance through depth maps, motion vectors, and camera parameters. 
Concurrently, video-to-video (V2V) editing methods focus on propagating edits across frames while preserving the features of the source video~\cite{tokenflow2024, fatezero2023, codef2024, pix2video2023, tuneavideo2023, text2videozero2023, cove2024}. Many such approaches leverage DDIM inversion to facilitate appearance modifications~\cite{i2vedit2024, magicedit2023, stablevideo2023}. However, these methods are fundamentally designed for local appearance changes; they struggle with non-local motion edits where the structural correspondence between frames is disrupted. When motion patterns are altered, the temporal alignment assumptions underlying these inversion-based approaches are violated.

\subsection{Motion-Guided Video Generation}

Motion control has emerged as a critical research direction, broadly categorized into trajectory-based and optical-flow-based methods. Trajectory-based approaches condition generation on point trajectories~\cite{tora2024, dragnuwa2023, draganything2024, dragavideo2023, imageconductor2024, boximator2024, i2vcontrol2024, 3dtrajmaster2024, flextraj2024, freetraj2024, trailblazer2024}, granting precise control over object paths, camera movement, and complex interactions. Conversely, optical flow-based methods~\cite{onlyflow2024, animateanything2024} utilize dense correspondence priors derived from optical flow estimators and point trackers~\cite{raft2020, tapir2023, bootstap2024, cotracker3_2024} to achieve fine-grained motion transfer.

Despite their impressive capabilities, these methods operate primarily as \textit{generators} rather than editors. Instead of modifying an input video directly, they extract attributes (e.g., optical flow) to condition the synthesis of an entirely new video. Recent trajectory-based methods~\cite{motionprompt2024, gowiththeflow2025, ati} attempt to bridge this by conditioning on single images and motion trajectories. However, while powerful for content creation, they fail to preserve the unrevealed visual context of existing videos when motion is modified. First-frame preserving methods like ReVideo~\cite{revideo2024} attempt to address this via inpainting but degrade when camera motion reveals content absent from the initial frame.

Our method addresses these limitations to enable true video-to-video motion editing. Specifically, we allow for flexible modification of object and camera trajectories while rigorously preserving the remaining video content. This approach generalizes effectively to arbitrary objects, diverse camera motions, and complex multi-element scenes.

\newcommand{\Nblobs}{N}                  % number of tracking blobs
\newcommand{\Hrgb}{H_{\text{rgb}}}       % RGB height
\newcommand{\Wrgb}{W_{\text{rgb}}}       % RGB width
\newcommand{\Hlat}{H_{\text{latent}}}       % latent height  
\newcommand{\Wlat}{W_{\text{latent}}}       % latent width
\newcommand{\Clat}{C_{\text{latent}}}       % latent channels

\newcommand{\Fframes}{F}                 % number of frames (model output)
\newcommand{\Flat}{F_{latent}}                 % number of frames (model output)
\newcommand{\Vfull}{V_{\text{full}}}     % full-length source video
\newcommand{\Ffull}{F_{\text{full}}}     % full video length in frames
\newcommand{\fstart}{f_{\text{start}}}   % start frame index
\newcommand{\fend}{f_{\text{end}}}       % end frame index
\newcommand{\fstartcf}{f_{\text{start}}^{\text{cf}}} % counterfactual start frame
\newcommand{\fendcf}{f_{\text{end}}^{\text{cf}}}     % counterfactual end frame

% Video variables
\newcommand{\Vcf}{V_{\text{cf}}}         % counterfactual video
\newcommand{\Vtarget}{V_{\text{target}}} % target video
\newcommand{\Vreal}{V_{\text{real}}}     % real video (alias for target)

% Track variables
\newcommand{\Tcf}{T_{\text{cf}}}         % counterfactual tracks
\newcommand{\Ttarget}{T_{\text{target}}} % target tracks
\newcommand{\Treal}{T_{\text{real}}}     % real tracks (alias for target)
\newcommand{\Tinit}{P_{\text{init}}}     % initial tracking points

% Blob rendering variables
\newcommand{\Bcf}{B_{\text{cf}}}         % counterfactual motion blobs
\newcommand{\Btarget}{B_{\text{target}}} % target motion blobs
% Method Section
\section{Our Approach}
% In this section we present our approach to develop a video-to-video motion editing tool composed of a mechanism to describe motion and a video diffusion model. This V2V motion editing tool allows users to change object movement while explicitly preserving static elements (e.g., moving a dog while keeping the background fixed); simultaneously manipulate object motion and camera perspective (e.g., panning the camera while an object moves); exercise temporal control (e.g., having the dog appear on second 5 instead of second 2); and apply these changes across any span of frames. We name these capabilities: object motion, camera control, time control of trajectories, and arbitrary frame specification.

% We demonstrate that a system enabling users to define ``motion edits''—which explicitly describe the desired change in motion—can successfully support these four capabilities. In this section, we first describe these capabilities and then explain the innovations that allow us to accomplish them, such as our proposed motion counterfactual video generation method and our video-to-video motion architecture.

In this section, we present a video-to-video motion editing framework that integrates a motion description mechanism with a video diffusion model. Our approach enables four core capabilities: object motion (altering movement while preserving static backgrounds, e.g., moving a dog but not the scene); camera control (simultaneously manipulating object and camera perspective, e.g., panning while an object moves); temporal control (adjusting trajectory timing, e.g., delaying an action to the 5th second); and arbitrary frame specification (applying edits across any frame span).

We demonstrate that explicitly defining `motion edits', which explicitly describe the desired change in motion, enables our system to robustly support these tasks. We first outline these capabilities in detail, followed by our key technical contributions: the motion counterfactual video generation method and our specialized video-to-video architecture.

\subsection{Editing Video through Motion Edits}

\begin{figure*}[t]
\centering
\includegraphics[width=1\linewidth]{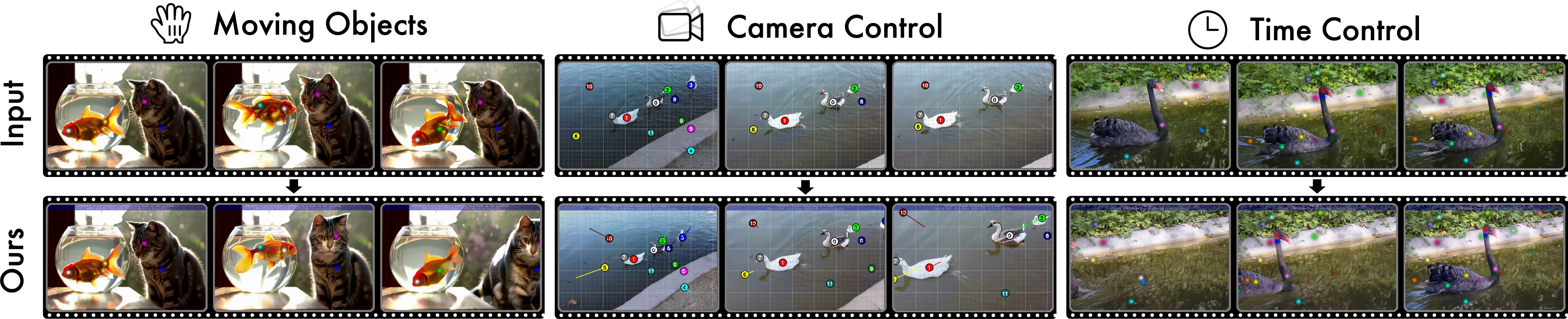}
\caption{From left to right respectively, \textbf{Cat Fish.} In the edited video, the cat moves away from the bowl. \textbf{Camera control.} In the edited video, the first frame is zoomed out, middle frame is identical, the last frame is zoomed in. \textbf{Duck Zoom.} The edited video exhibits different content for a given frame (time) than the original, e.g. in the edited video, the duck is not visible in the first frame whereas it is visible in the original.}
\vspace{-15pt}
\label{fig:three_strips}
\end{figure*}

\textbf{Moving Objects} By identifying an object's trajectory and editing it, we can change the motion of the object as the video progresses. As shown in Figure~\ref{fig:teaser} and ~\ref{fig:three_strips}, this can have high level effects such as changing the ultimate outcome of a scenario and is a flexible tool for many applications, such as re-timing subjects, improve video aesthetics by moving occluders, or recomposing a video and its parts in motion. 
% While a subset of these capabilities have been tackled before, unlike prior methods, we present a general approach that works in the video-to-video scenario and on any object, including those that are not present in the first frame of the video.

\noindent\textbf{Camera Control} With our motion editing scheme we can control camera pose and motion in the video relative to the scene. We estimate a dynamic pointmap~\cite{zhang2024monst3r}, reproject it into each frame using user-specified camera extrinsics and intrinsics, and then solve for deviations in the pointwise trajectories. This allows us to dynamically change the position and focal length of the camera in any frame while also preserving the video content. We show this in the Swan example in Figure~\ref{fig:teaser} where the Swan is swimming and the ripples in the water are preserved despite changes in the camera position and in~\ref{fig:three_strips} where each frame has a different zoom level with same scene content.

\noindent\textbf{Time Control} Our method allows users to control trajectories of specific elements in a video independent of the global timeline. This enables delaying or accelerating an object's trajectory, such as making a subject appear on second 5 instead of second 2, while preserving the background's original motion. As shown in Figure~\ref{fig:three_strips}, we can delay the appearance of a duck until later frames, effectively decoupling the subject's timeline from that of the scene.

\begin{figure}[h]
\centering
\vspace{-4pt}
\includegraphics[width=\linewidth]{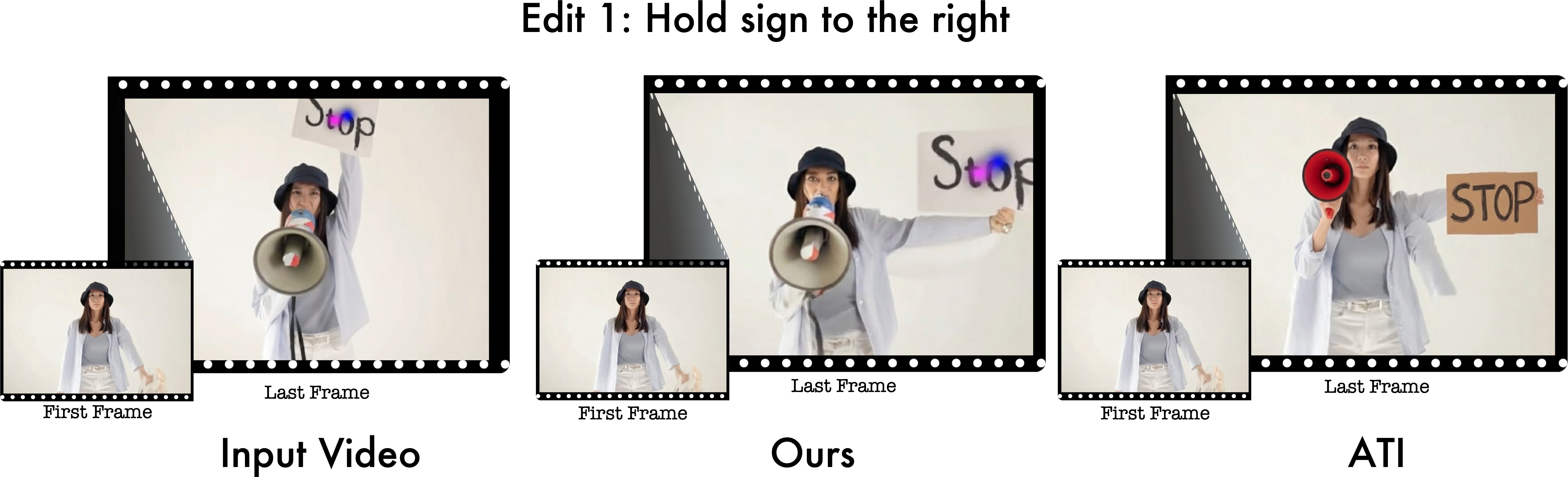}
\vspace{-20pt}
\caption{\textbf{Controlling Content on Any Frame.} By conditioning on the full video, we can move and preserve content appearing on any frame. Methods like ATI rely on the first frame, failing to control objects, like the sign, that emerge mid-sequence.}
\label{fig:anyframe}
\vspace{-10pt}
\end{figure}

\noindent\textbf{Arbitrary Frame Specification} Unlike image-to-video approaches that rely on the first frame for content generation~\cite{gowiththeflow2025, motionprompt2024, revideo2024}, our framework supports editing objects that appear at any point in the video. Relying on the initial frame severely restricts possible edits and fails to account for elements that emerge later. Additionally, the motion of the rest of the video is entirely hallucinated whereas ours can conserve it in part or entirely. By conditioning on the full video, we enable precise control over mid-stream objects, such as the stop sign in Figure~\ref{fig:anyframe}. 

%\begin{figure}[h]
%    \centering
%    \includegraphics[width=\linewidth]{\figfolder/TestPrep}
%    \caption{Video preparation process for training data generation. A dataset video is split to create input and target video pairs with shared tracking points.}
%    \label{fig:video_prep}
%\end{figure}
%% \TODO{Change figure.}

\begin{figure}[h]
    \centering
    \includegraphics[width=\linewidth]{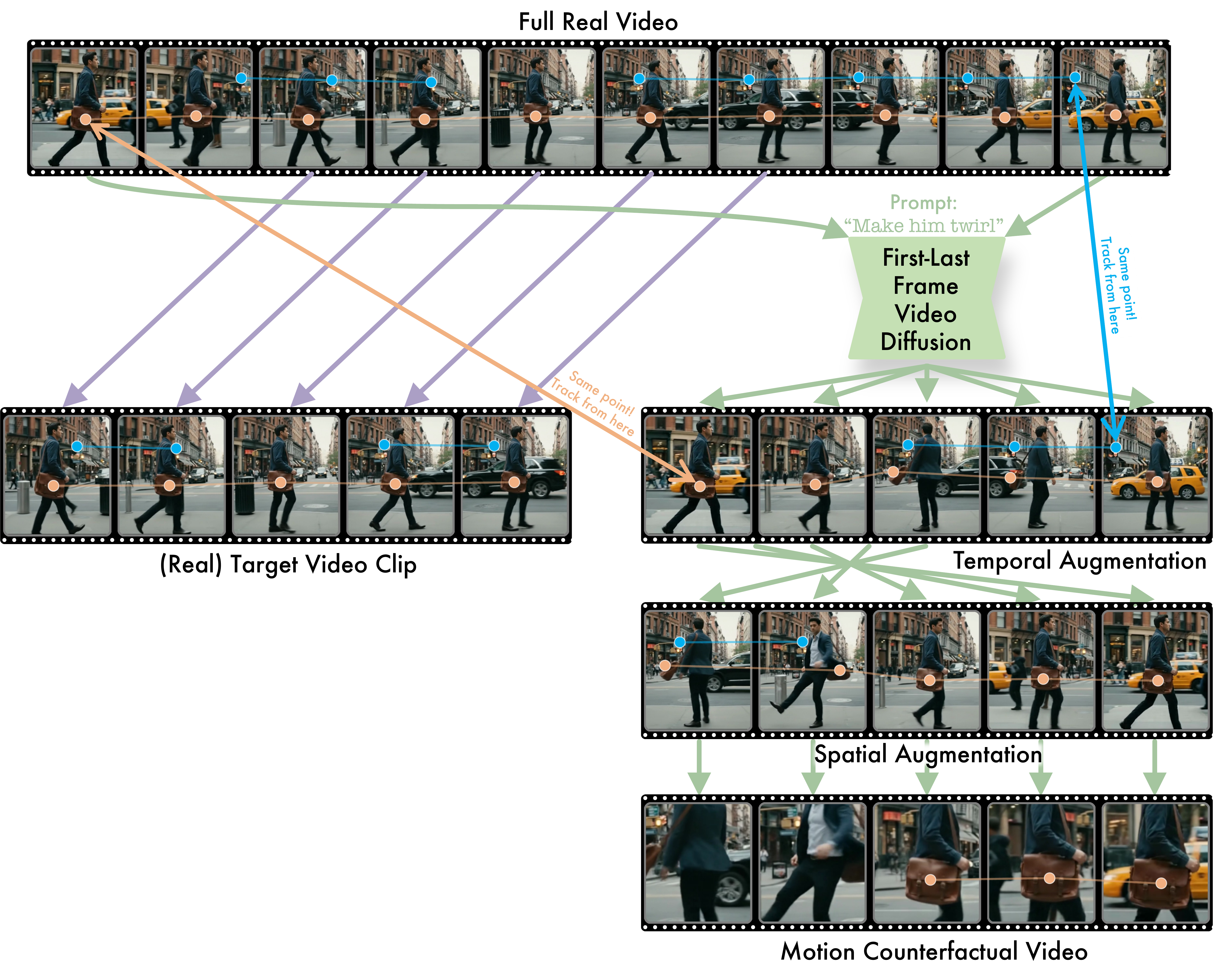}
    \caption{\textbf{Counterfactual data generation process.} In order to generate a real / counterfactual video pair and its corresponding trajectories, we take a full real video, extract a video clip, then create a counterfactual video. The counterfactual has new motion from the video generator, as well as temporal and spatial augmentations. In order to ensure we have two corresponding set of tracks, we specifically use the first and last frames, which directly match the original video, to anchor the tracks for the counterfactual.}
    \label{fig:video_prep}
\end{figure}
% \TODO{Change figure.}

\begin{figure*}[t]
    \centering
    \includegraphics[width=1\linewidth]{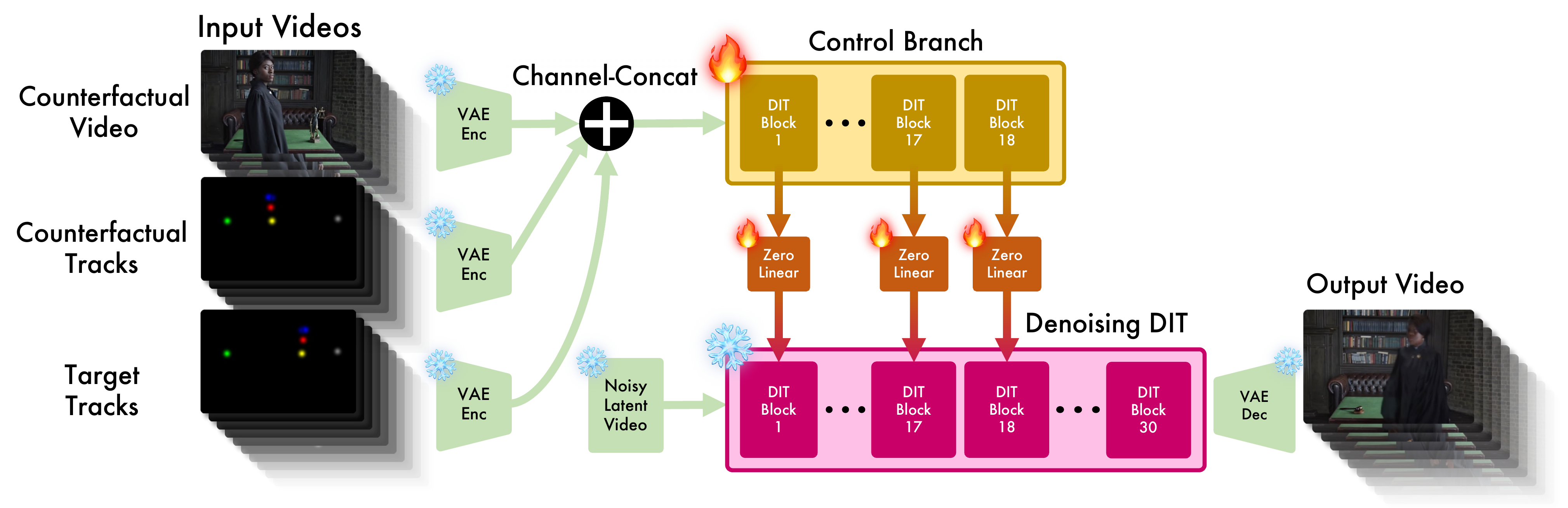}
    \caption{\textbf{Our motion-conditioned video diffusion architecture.} We extend a T2V DiT model with a control branch that processes three additional video conditioning channels: the counterfactual video, counterfactual motion tracks, and target motion tracks. The control branch duplicates the first 18 transformer blocks and integrates with the main branch through zero-initialized MLPs, similar to ControlNet.}
    \label{fig:architecture}
\end{figure*}

\subsection{Motion Counterfactual Video Generation}
Our approach requires training data consisting of video pairs with identical visual content but different motion patterns. We generate these \textit{motion counterfactual videos} $\Vcf$ and corresponding \textit{target videos} $\Vtarget$ from raw videos using a systematic process that ensures trackable point correspondences between video pairs (Figure~\ref{fig:video_prep}).

Given a source video $\Vfull$ of length $\Ffull$ frames, we generate video pairs as follows. First, we extract the target video $\Vtarget$ by selecting a contiguous frame chunk of length $\Fframes$ with random starting frame $\fstart \sim \text{Uniform}(0, \Ffull - \Fframes)$. We keep real video as targets to ensure the model trains toward realistic motion and appearance.

For the counterfactual video $\Vcf$, we randomly select start and end frame indices $\fstartcf, \fendcf \sim \text{Uniform}(0, \Ffull-1)$ and choose one of two generation strategies:

\noindent \textbf{Frame Interpolation:} We use a video diffusion model conditioned on frames $\fstartcf$ and $\fendcf$ to generate a $\Fframes$-frame video. This adds new content via LLM-generated prompts—e.g., instructing a walking person to ``twirl'' (Figure~\ref{fig:video_prep}). This provides more data than first-frame-only methods, allowing the model to use more of the input video.

\noindent \textbf{Temporal Resampling:} We extract $\Fframes$ frames evenly spaced between $\fstartcf$ and $\fendcf$ from $\Vfull$. This creates natural speed variations, temporal shifts, and sometimes reversed motion when $\fstartcf > \fendcf$.

Next, we establish point correspondences between the video pair. We initialize $\Nblobs \sim \text{Uniform}(1, 64)$ tracking points with coordinates $(t_i, x_i, y_i)$ where $x_i$ and $y_i$ sampled uniformly from frame dimensions $\Wrgb, \Hrgb$ respectively. For temporal coordinates $t_i$:
\begin{itemize}
    \item \textbf{Temporal resampling:} Frame indices are sampled from frames present in both $\Vtarget$ and $\Vcf$
    \item \textbf{Frame interpolation:} Frame indices are restricted to $\{\fstartcf, \fendcf\}$ to ensure correspondence
\end{itemize}

We use TAPNext~\cite{tapnext}, a bidirectional point tracker, on $\Vfull$ with these initial points to obtain target tracks $\Ttarget$. For counterfactual tracks $\Tcf$: in temporal resampling cases, we use the same tracker output; for interpolation cases, we first replace the corresponding frames in $\Vfull$ with the interpolated $\Vcf$ frames before running TAPNext~\cite{tapnext}. 
%\fcole{Do we check if the trajectory tracking worked for the generated counterfactuals, or filter for bad outputs? If so what is the success rate?}

Finally, we apply geometric augmentations to the counterfactual videos including random sliding crops, rotations, and scale changes, with the same transformations applied to the corresponding tracking points to maintain correspondence. These artificial moving crops approximate multi-view videos and ensure perfect temporal synchronization—giving the model a bias toward synchronizing appearance when otherwise unspecified.
\paragraph{Trajectory Representation}
% \fcole{Trajectory representation seems important and maybe should be its own subsection}
A key part of our method is our representation of motion throughout videos. Our model is conditioned on three videos: the counterfactual video $\Vcf$, the rendered counterfactual motion tracks $\Bcf$, and the rendered target motion tracks $\Btarget$, each of dimension $\mathbb{R}^{\Fframes \times 3 \times \Hrgb \times \Wrgb}$. Additionally, like our base model, text prompts provide semantic conditioning, which we will leave out of equations in this section for brevity.

We rasterize the tracking information as colored Gaussian blobs on black backgrounds to create motion conditioning channels. For each training sample, we randomly select $\Nblobs$ distinct random colors. Each tracking point is rendered as a Gaussian blob with standard deviation of 10 pixels in its assigned color in both the counterfactual tracks video $\Bcf$ and target tracks video $\Btarget$, with blobs only drawn when the corresponding point is visible (not occluded) as reported by the point tracker. We also tried representations similar to \citep{das2025}, but found that both large number of points and the lack of distinct colors made it a weaker control signal.
%\todo{Add interesting facts on other things we tried.}

The tracks are subject to dropout during training, with target motion blobs $\Ttarget$ experiencing higher dropout rates than conditioning tracks $\Tcf$ to improve robustness and prevent overfitting to specific motion patterns. During inference, we limit the number of point correspondences to approximately 20, as the model fails to follow all correspondences when given too many points.

\subsection{Model Architecture}
We use a pre-trained T2V DiT as our base model~\cite{cogvideox2024}. In order to condition on motion and input videos we incorporate a control branch duplicating the first 18 transformer blocks of the DiT that feeds into the main branch using zero-initialized MLPs. Conceptually the control branch is similar to a ControlNet~\cite{controlnet2023} applied to a DiT architecture. We are inspired by the architecture proposed in DiffusionAsShader~\cite{das2025}, but our implementation has the key difference of allowing conditioning on three video tracks and using a control branch patchifier that handles $48 = 3 \times 16$ input channels for the three conditioning videos in latent space.

The control branch tokens are fed through zero-init~\cite{controlnet2023}, channel-wise MLPs and then added to the main branch token values in their respective transformer blocks. The base model processes the noisy video being denoised along with text conditioning, while our control branch handles the three additional video conditioning channels $\Vcf, \Bcf, \Btarget$. All video inputs are encoded using a 3D Causal VAE~\cite{cogvideox2024}, which compresses RGB videos of shape $\Fframes \times 3 \times \Hrgb \times \Wrgb$ to latent representations of shape $\Flat \times \Clat \times \Hlat \times \Wlat$ where $\Clat = 16, \Flat=\left(\frac{\Fframes-1}{4}+1\right), \Wlat=\frac{\Wrgb}{8}, \Hlat=\frac{\Hrgb}{8}$. The main branch is frozen while the control branch is trained.

During training, the model learns to generate target videos $\Vtarget$ that follow specified motion patterns and satisfy the given correspondences between counterfactual and target tracks. The training objective is conditioned on the counterfactual video $\Vcf$, its tracks $\Tcf$, the target tracks $\Ttarget$, and a text prompt describing the scene. This formulation successfully teaches the model to transfer motion patterns from the target tracks while maintaining the visual realism of target video content.

The task we tackle is harder than a typical ControlNet task where the structure is usually given to the model. For example an edge-to-image ControlNet has a good idea of what the structure of the output should be with edges as input. Surprisingly, our adapter works despite the inputs (video + motion blobs) lacking spatiotemporal synchronization with the output. We hypothesize that transformer blocks do non-trivial work to achieve this capability. 

%We tried different architectural solution such as finetuning the diffusion model with extra control channels but obtained underwhelming results. \fcole{last sentence is confusing: was this different architectural solution superseded by the current solution, or was it an attempt to improve the current approach further? Also not clear what exactly the extra control channels were. Maybe just delete this?}

% Experiments Section
% User study data - single source of truth (DRY!)
% Study parameters
\newcommand{\NumParticipants}{41}

% Win rates
\newcommand{\OursQOne}{70}
\newcommand{\ATIQOne}{24}
\newcommand{\ReVideoQOne}{1}
\newcommand{\GWTFQOne}{5}

\newcommand{\OursQTwo}{71}
\newcommand{\ATIQTwo}{24}
\newcommand{\ReVideoQTwo}{2}
\newcommand{\GWTFQTwo}{3}

\newcommand{\OursQThree}{69}
\newcommand{\ATIQThree}{25}
\newcommand{\ReVideoQThree}{1}
\newcommand{\GWTFQThree}{5}

% Helper to convert percentage to decimal for plots
\newcommand{\pct}[1]{\fpeval{#1/100}}

\section{Results}
\label{sec:experiments}

\subsection{Implementation Details}

We use CogVideoX-5B~\cite{cogvideox2024} as our base text-to-video model for both the finetuned counterfactual video generation model and the V2V editing model.
Training was conducted on 8 H100 GPUs for one week using standard latent diffusion training with L2 loss.
We set $\Fframes = 49$,
with input resolution of $480 \times 720$ pixels,
corresponding to latent dimensions of $60 \times 90$.
We use $\Nblobs$ varying between 1 and 64 during training and set the control branch depth appropriately.
We use a learning rate of $10^{-4}$ and a dataset size of $100,000$ videos,
for $15,000$ iterations with an effective batch size of $32$.
We use an internal video dataset with 500,000 samples.
\newcommand{\Vtest}{V_{\text{test}}}     % test video
\newcommand{\Vzero}{V_0}                 % first half of test video  
\newcommand{\Vone}{V_1}                  % second half of test video
\newcommand{\Voneprime}{V_1'}            % reversed second half
\newcommand{\Ntest}{N_{\text{test}}}     % number of test videos
\newcommand{\Npoints}{N_{\text{points}}} % number of tracking points for evaluation

We evaluate our motion editing approach through user studies and quantitative metrics, comparing against state-of-the-art motion control methods. %Details of how these baselines are implemented can be found in the appendix.

%\TODO{Explain how each baseline algorithm (ATI, ReVideo, Go-with-the-Flow, MotionPrompting) was implemented to work with our evaluation framework and track inputs. ---- PUT THAT IN APPENDIX }

\subsection{User Study}

We conducted a user study comparing our method against three baselines: ATI~\cite{ati},
a trajectory-guided image-to-video method based on WAN 2.1~\cite{wan};
ReVideo~\cite{revideo2024};
and Go-with-the-Flow (GWTF)~\cite{gowiththeflow2025}.
We manually created 20 test videos spanning diverse scenarios including object motion editing,
camera motion changes,
and complex scenes with multiple moving elements. \NumParticipants{} participants compared all four methods using the interface shown in the Supplementals,
selecting the best video for each of three questions per test case:
\begin{itemize}
    \item \textbf{Q1:} ``Which video better preserves the input video's content?''
    \item \textbf{Q2:} ``Which video better reflects the desired motion?''
    \item \textbf{Q3:} ``Which video is overall a better edit of the input video?''
\end{itemize}

\begin{table}[h]
\centering
\small
\begin{tabular}{l|cccc}
\hline
Question & Ours & ATI & ReVideo & GWTF \\
\hline
Q1: Content ($\uparrow$) & \textbf{\OursQOne\%} & \ATIQOne\% & \ReVideoQOne\% & \GWTFQOne\% \\
Q2: Motion ($\uparrow$) & \textbf{\OursQTwo\%} & \ATIQTwo\% & \ReVideoQTwo\% & \GWTFQTwo\% \\
Q3: Overall ($\uparrow$) & \textbf{\OursQThree\%} & \ATIQThree\% & \ReVideoQThree\% & \GWTFQThree\% \\
\hline
\end{tabular}
\caption{\textbf{User study win rates across all methods.}
Participants selected the best video for each question.
Our method consistently wins across all evaluation criteria.}
\label{tab:user_study}
\end{table}

Table~\ref{tab:user_study} show that users consistently ranked our method highest across all questions,
with win rates around 70\% compared to 25\% for ATI and less than 5\% for ReVideo and GWTF,
demonstrating superior content preservation and motion control.

\subsection{Quantitative Evaluation}

We developed a quantitative evaluation protocol using photometric reconstruction error to assess motion editing quality. 

\subsubsection{Dataset Construction}

We curated a dataset of $\Ntest = 100$ test videos using the following protocol.
Given a source video $\Vtest$ of length $\Ffull$ frames,
we split it temporally at the midpoint to obtain $\Vzero = \Vtest\left[1:\Ffull/2\right]$ and $\Vone = \Vtest\left[\Ffull/2:\Ffull\right]$.
We then create the counterfactual input by temporally reversing $\Vone$ to get $\Voneprime$,
ensuring temporal continuity between $\Vzero$ and $\Voneprime$ (i.e., the last frame of $\Vzero$ matches the first frame of $\Voneprime$).

We selected random internet videos not seen during training where significant content appears in middle frames but is not visible in the first frame.
To quantify this,
we initialize $\Npoints = 25$ tracking points at the temporal midpoint of each video and track them bidirectionally using TAPNext~\cite{tapnext}.
We retain only videos where a substantial number of points become occluded when tracked to both the first and last frames.

\begin{figure}[htb]
    \centering
    \includegraphics[width=\linewidth]{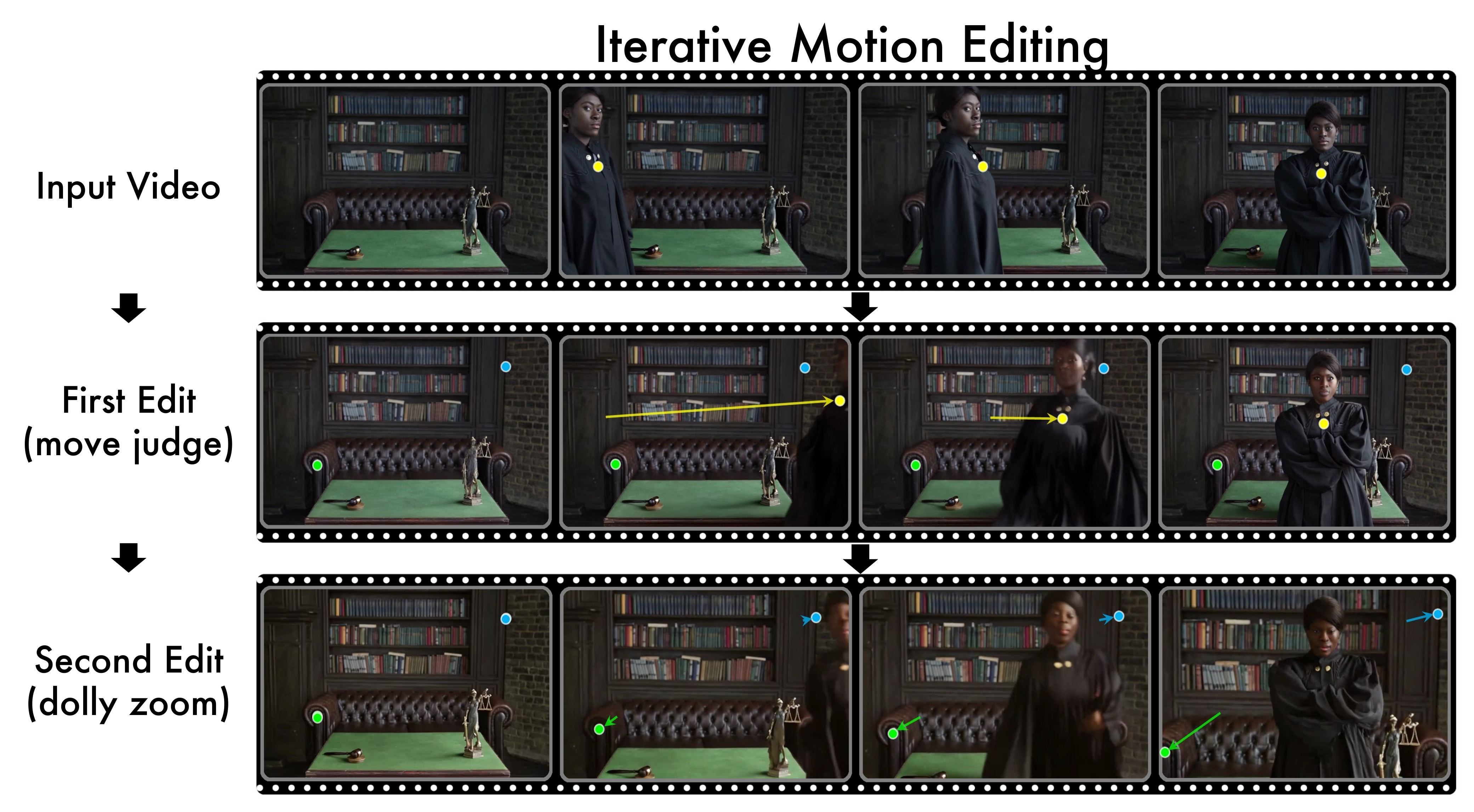}
    \caption{\textbf{Iterative editing.}
Outputs can become inputs for subsequent edits,
enabling complex sequential motion changes. Yellow dots used for first edit, green/cyan for second. Arrows added from old to new position for ease of visualization. }
    \label{fig:iterative_editing}
\end{figure}

\subsubsection{Evaluation Protocol}

For each test case,
we use $\Vzero$ as input and $\Vone$ as the target video.
We provide both our method and ATI with identical motion trajectories extracted from $\Vone$ and measure reconstruction quality using frame-wise L2 loss:

\begin{equation}
L_2 = \frac{1}{\Fframes} \sum_{i=1}^{\Fframes} \|I_i^{\text{pred}} - I_i^{\text{target}}\|_2^2
\end{equation}

where $\Fframes$ is the number of frames,
$I_i^{\text{pred}}$ is the $i$-th predicted frame,
and $I_i^{\text{target}}$ is the corresponding target frame.

\begin{table}[h]
\centering
\begin{tabular}{l|c|c|c}
\hline
\textbf{Method} & $\mathbf{L_2}\,(\downarrow)$ & \textbf{SSIM}\,($\uparrow$) & \textbf{LPIPS}\,($\downarrow$) \\
\hline
Ours               & \textbf{0.024} & \textbf{0.098} & \textbf{0.031} \\
ATI                & 0.038 & 0.094 & 0.072 \\
Go-with-the-Flow   & 0.067 & 0.089 & 0.088 \\
ReVideo            & 0.096 & 0.080 & 0.106 \\
\hline
\end{tabular}

\caption{Evaluation of photometric reconstruction error for our method and ATR. Our method achieves significantly lower L2 reconstruction error.}
\label{tab:quantitative}
\vspace{-10pt}
\end{table}

Our method achieves substantially lower reconstruction error (Table~\ref{tab:quantitative}),
confirming that our full-video approach better preserves content compared to first-frame generation methods,
particularly in scenarios involving content not visible in initial frames.

\subsection{Qualitative Comparisons}

\noindent \textbf{Iterative Edits.}
One of the strengths of our technique is that it can be applied iteratively - taking the output of one run and using it as input for a successive video edit. This allows users to chain multiple simple, intuitive edits together in order to achieve a very complicated edit. This iterative editing also provides more immediate feedback to the user making the process more transparent and easier to control. In~\cref{fig:iterative_editing}, we show that a complex edit (an object motion and a camera change) can be decomposed into its core parts and applied one by one. While this example demonstrates some degree of subject drift, this can be attributed in part to the quality of the base video model. We believe that future versions of our method will be able to be applied infinitely.

\begin{figure*}[t]
    \centering
    \includegraphics[width=1\linewidth]{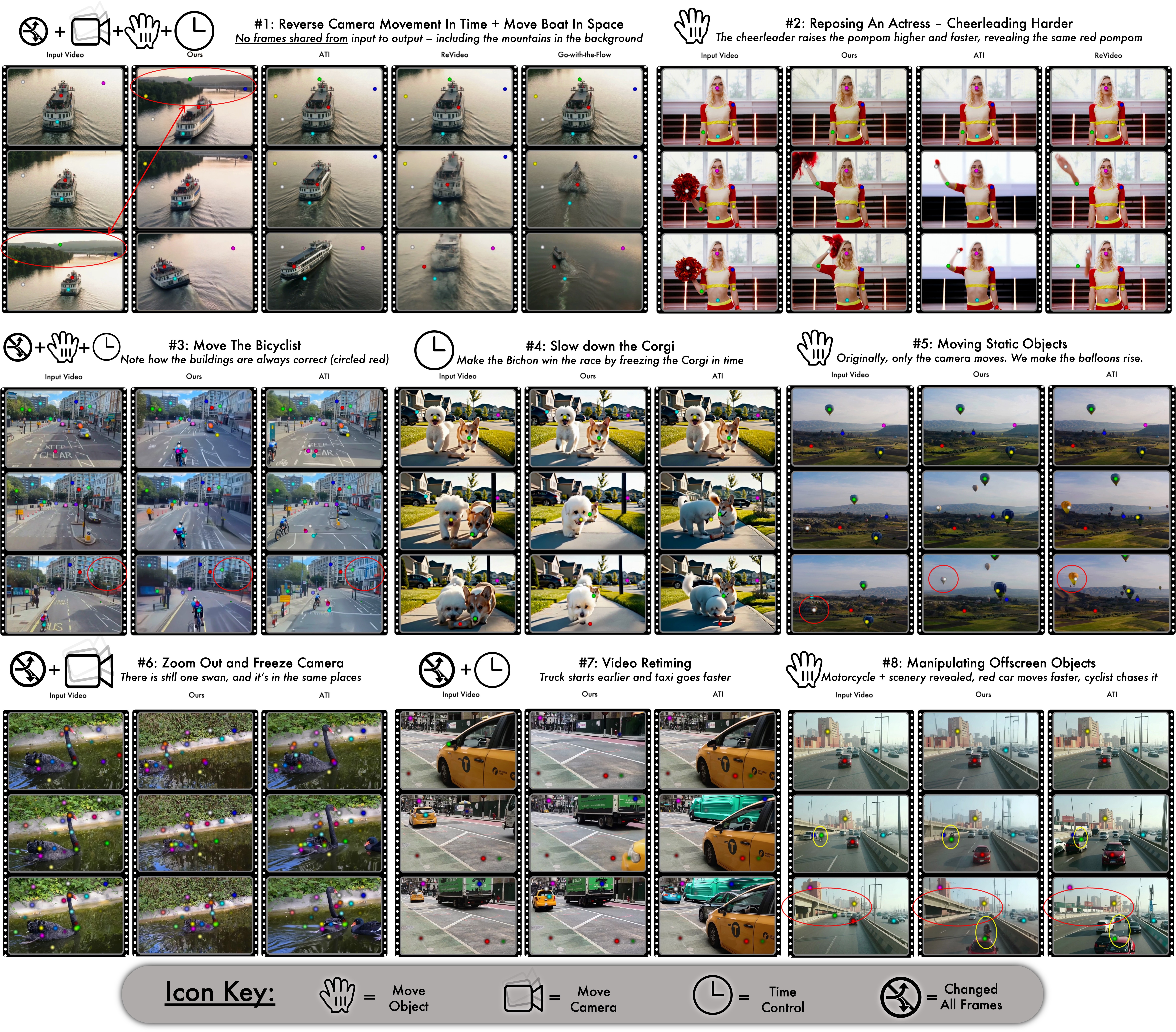}
    \caption{Comparison of our method vs.\ baselines across eight challenging motion editing scenarios.
Each row shows a different editing task with input video,
our result,
and ATI's result (with additional baselines shown for subfigure 4).
\textbf{Icon key:} Human Pose (modifying human motion),
Move Object (repositioning objects),
Move Camera (changing camera motion),
Time Control (retiming events),
Changed All Frames (no shared frames between input/output—impossible for image-to-video methods).
Colored dots track correspondence points throughout the video;
dot presence/absence indicates object visibility.
Red circles highlight key differences where baselines fail.}
    \label{fig:mega_comparison}
\end{figure*}

\noindent\textbf{Baseline comparisons.} Figure~\ref{fig:mega_comparison} compares our method against several baselines in multiple video editing scenarios, each of which demonstrate the capabilities of our motion edits. We primarily compare against ATI~\cite{ati},
a trajectory-guided image-to-video method based on WAN 2.1~\cite{wan},
our strongest baseline despite using a more powerful base model than our CogVideoX base.
Subfigure 4 additionally shows ReVideo~\cite{revideo2024} and Go-with-the-Flow~\cite{gowiththeflow2025},
which rated poorly in user evaluation—ReVideo lacks text conditioning and Go-with-the-Flow was not designed for point control.

\noindent\textbf{Edit \#1: Complex Edits on the Boat Scene.} This edit moves the boat left and shifts the camera so that mountains from the original's last frame appear in the edit's first. This requires specifying a substantial temporal trajectory change and holistic knowledge of the scene content. Ours is the only method that realistically moves the boat while correctly adjusting the camera to reveal the mountains at the beginning of the video.

\noindent\textbf{Edit \#2: Reposing a Cheerleader.} This edit raises the cheerleader's arms. The challenge involves preserving the red pom-pom, which is absent from the first frame. Ours successfully modifies the motion while retaining this content. In contrast, ATI and ReVideo rely solely on the first frame, leading to unnatural movements and a failure to preserve the pom-pom.

\noindent\textbf{Edit \#3: Move The Bicyclist.} This edit controls a cyclist visible only in the final frame of the original video. Ours correctly propagates the cyclist and tracking dots (cyan, magenta, white) throughout. ATI, lacking full temporal context, misplaces the cyclist and synthesizes wrong buildings (red circles).

\noindent\textbf{Edit \#4: Dog Race.}
Differential timing breaks single-frame-based methods. We decelerate the Corgi (green dot) to reverse the race outcome while keeping the Bichon steady. This requires independent temporal control; ATI fails to decouple the motions, incorrectly copying the Bichon and transforming a light pole into a tree.

\noindent\textbf{Edit \#5: Moving Static Balloons.}
We add upward motion to stationary balloons. The white balloon (white dot), which appears mid-video, challenges partial information methods. While ATI moves visible balloons, it renders the initially hidden white balloon orange due to missing appearance data. Our method uses full video context to maintain correct colors.

\noindent\textbf{Edit \#6: Zooming out on the Swan} 
In this DAVIS~\cite{davis2016} example, we transform a panning shot into a static, zoomed-out view. The output field of view differs entirely from the input, yet the swan must remain anchored to specific vegetation. Lacking full spatial context, ATI synthesizes a second swan and produces inconsistent motion.

\noindent\textbf{Edit \#7: Retiming a taxi.}
We do a complex isolated retiming of taxi and truck movement. This requires complete temporal understanding; ATI's single-frame generation cannot achieve this reversal. Figure~\ref{fig:mega_comparison} compares our method against ATI (WAN 2.1-based), as well as ReVideo and Go-with-the-Flow (in Subfigure 4), both of which were rated poorly due to their design limitations.

\noindent \textbf{Edit \#8: Moving an Offscreen Car.}
As the camera follows a red car, a motorcyclist enters late. We reposition this initially invisible rider behind the car while maintaining consistent background architecture. Lacking future frames to reference the rider and buildings (red circles), ATI synthesizes incorrect content.

\noindent\textbf{Discussion.}
These scenarios highlight I2V limitations: conditioning only on the first frame prevents leveraging information from the full input. Our V2V formulation enables bidirectional flow, allowing outputs to pull content from \emph{any} input frame. This handles offscreen content, camera changes, and reordering—challenges where I2V methods like ReVideo~\cite{revideo2024}, Go-with-the-Flow~\cite{gowiththeflow2025}, and MotionPrompting~\cite{motionprompt2024} fail.

\section{Conclusion}

We developed a new video-to-video motion editing algorithm that allows us to edit the motion of objects, subjects and camera pose in user-provided videos. To the best of our knowledge it is the first in its class, compared to other work that control motion in the image-to-video setup. Our algorithm has a comfortable user interface, where a user drags sparse point trajectories to control objects or camera motion.

\clearpage
{
    \small
    \bibliographystyle{ieeenat_fullname}
    \bibliography{main}
}

\clearpage
\setcounter{page}{1}
\maketitlesupplementary

\section{Human Interaction}

\begin{figure}[h]
    \centering
    \includegraphics[width=1\linewidth]{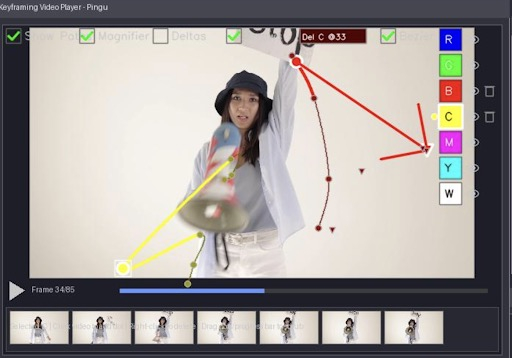}
    \caption{\textbf{Interface for creating motion edits.} The red arrow shows the transformation from source trajectory (line) to target trajectory (triangle).}
    \label{fig:motion_editor_gui}
\end{figure}

Our motion editing interface (Figure~\ref{fig:motion_editor_gui}) provides an intuitive way to specify complex motion changes without requiring any laborious segmentation or rotoscoping. Users simply click points on the video to initialize points which are then tracked bidirectinally to create source trajectories. Then, the user manipulates these trajectories using Bezier splines to define the desired target motion, with arrows indicating the transformation from source to target.

The red arrow in the figure illustrates a typical edit, showing the transformation from the source trajectory (line) to the target trajectory (triangle). The interface allows users to scrub through video frames and place trajectory points as needed. Different tracking points are represented by different colors: red, green, blue, cyan, magenta, yellow, and white.

\section{User Study}

As described in Section~\ref{sec:experiments}, we conducted a user study with \NumParticipants{} participants evaluating 20 test videos to compare our method against state-of-the-art baselines.

\begin{figure}[h]
    \centering
    \includegraphics[width=1\linewidth]{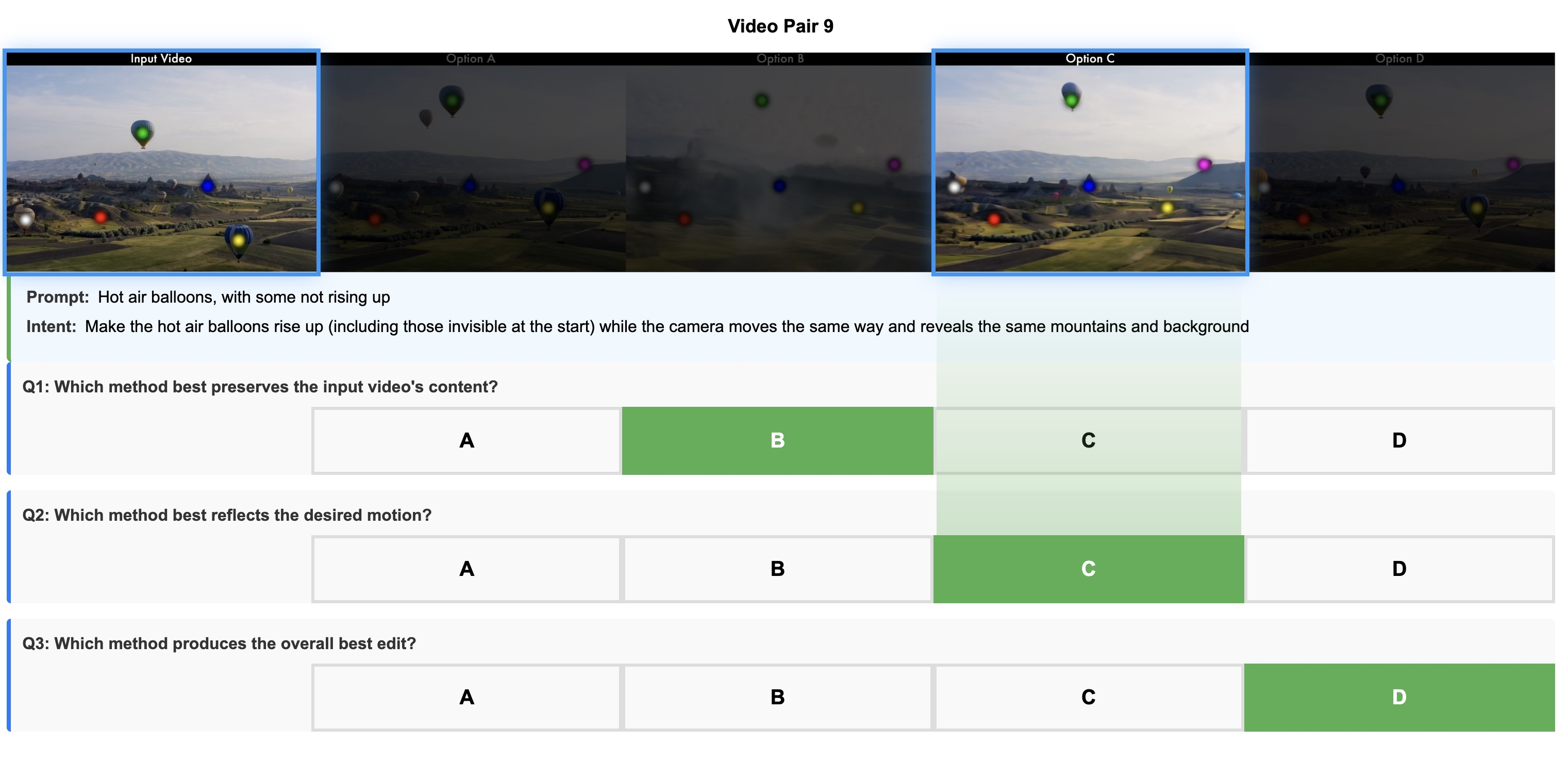}
    \caption{\textbf{User study evaluation interface.} Side-by-side video comparisons with visualized motion edits.}
    \label{fig:user_study_interface}
\end{figure}

\paragraph{Evaluation Protocol}
Participants used the interface shown in Figure~\ref{fig:user_study_interface} to compare our method against three baselines: ATI~\cite{ati}, ReVideo~\cite{revideo2024}, and Go-with-the-Flow~\cite{gowiththeflow2025}. The interface presents side-by-side comparisons of the original input video alongside results from our method and all baselines, with colored tracking dots visualizing the motion edits so users can clearly see how each method interprets the desired motion changes. For each test case, participants were asked three questions:

\begin{itemize}
    \item \textbf{Q1:} ``Which video better preserves the input video's content?''
    \item \textbf{Q2:} ``Which video better reflects the desired motion?''
    \item \textbf{Q3:} ``Which video is overall a better edit of the input video?''
\end{itemize}

% User study win rates graph
\begin{figure}[h]
  \centering
  \begin{tikzpicture}
    \begin{axis}[
      ybar,
      bar width=6pt,
      width=\columnwidth,
      height=5cm,
      ymin=0, ymax=1,
      ylabel={Win Rate},
      symbolic x coords={Ours, ATI, ReVideo, GWTF},
      xtick=data,
      ymajorgrids,
      grid style={dashed,gray!40},
      enlarge x limits=0.15,
      legend style={
          at={(0.98,0.98)},
          anchor=north east,
          draw=black,
          font=\footnotesize,
          fill=white,
          cells={anchor=west}
      },
      legend image code/.code={
          \draw[#1] (0cm,-0.1cm) rectangle (0.3cm,0.1cm);
      },
      tick label style={font=\footnotesize},
      label style={font=\footnotesize},
      title style={font=\footnotesize},
      title={User Study Win Rates}
    ]

    % Solid color bars using data macros with \pct helper
    \addplot[fill=blue!60, draw=black] coordinates {
      (Ours,\pct{\OursQOne})
      (ATI,\pct{\ATIQOne})
      (ReVideo,\pct{\ReVideoQOne})
      (GWTF,\pct{\GWTFQOne})
    };
    \addplot[fill=red!60, draw=black] coordinates {
      (Ours,\pct{\OursQTwo})
      (ATI,\pct{\ATIQTwo})
      (ReVideo,\pct{\ReVideoQTwo})
      (GWTF,\pct{\GWTFQTwo})
    };
    \addplot[fill=brown!60, draw=black] coordinates {
      (Ours,\pct{\OursQThree})
      (ATI,\pct{\ATIQThree})
      (ReVideo,\pct{\ReVideoQThree})
      (GWTF,\pct{\GWTFQThree})
    };

    \legend{Q1: Content,Q2: Motion,Q3: Overall}
    \end{axis}
  \end{tikzpicture}
  \caption{User study win rates per question (see Table~\ref{tab:user_study} for values).}
  \label{fig:winrates-by-question}
\end{figure}
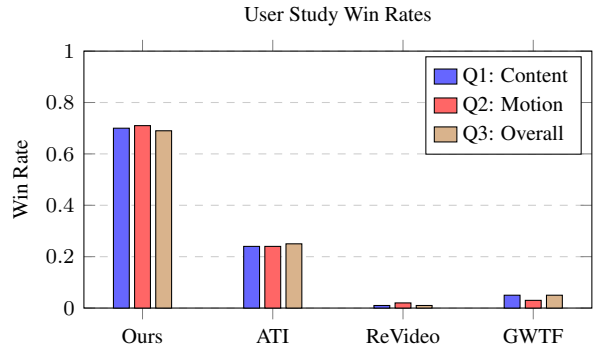

The results, visualized in Figure~\ref{fig:winrates-by-question} and detailed in Table~\ref{tab:user_study}, show users greatly preferred our method, with rates around 70\% compared to approximately 25\% for ATI and less than 5\% for ReVideo and Go-with-the-Flow.

\section{Quantitative Evaluation Dataset}

\begin{figure}[h]
    \centering
    \includegraphics[width=\linewidth]{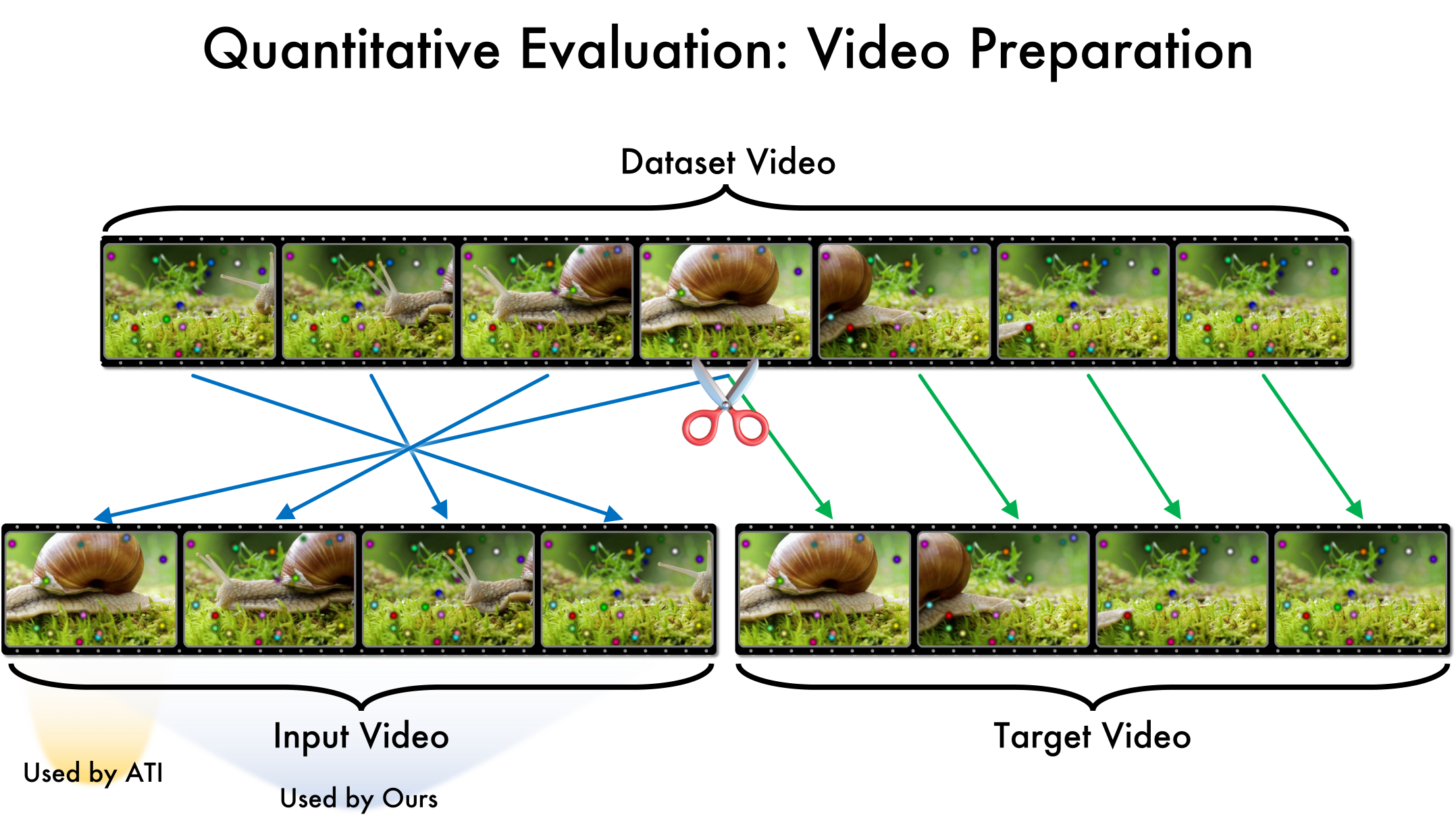}
    \caption{\textbf{Test Data Generation.} A video is separated at the middle, and then one half is reversed. This results in two videos with a common starting frame.}
    \label{fig:video_prep_suppl}
\end{figure}

Our quantitative evaluation dataset construction is fully described in Section~\ref{sec:experiments} (Dataset Construction subsection). We created $\Ntest = 100$ test videos by splitting videos at their temporal midpoint and reversing one half to create video pairs with common starting frames (Figure~\ref{fig:video_prep_suppl}). This protocol ensures that image-to-video baselines, which require first-frame correspondences, receive inputs they can properly handle since the tracking points match the first frame. The dataset specifically includes videos where significant content appears in middle frames but not in the first frame, quantified by tracking $\Npoints = 25$ points bidirectionally from the midpoint.

\section{Baseline Implementation Details}

All baseline methods are image-to-video (I2V) algorithms with motion control, fundamentally different from our video-to-video (V2V) approach:

\paragraph{ATI~\cite{ati}}
ATI is based on Wan2.1. It is a point-based image-to-video algorithm with controllable motion. For our baseline, we take the target tracks and apply it to the first frame of our counterfactual videos, using the target prompts for text guidance. We use the default number of diffusion steps and CFG as provided by their public code repository.

\paragraph{ReVideo~\cite{revideo2024}}
ReVideo is based on Stable Video Diffusion. It takes no prompt as an input. It is an image-to-video point-based motion-controllable algorithm, with the ability to specify editable regions. Since we are avoiding manual labor such as rotoscoping we designate the entire video as an editable region. 

\paragraph{Go-with-the-Flow~\cite{gowiththeflow2025}}
Go-with-the-Flow is not a point-based motion control algorithm, but is instead an image-to-video motion-controllable algorithm driven by warped noise, which often comes from optical flow. To make this baseline work, we run the raterized target tracks through RAFT to get optical flows, and from that create warped noise that is used to generate the output videos. We use a more recent Wan2.2-based version of Go-with-the-Flow to test with, as it is a tougher baseline than their original CogVideoX-5B model.

The fundamental limitation of all these baselines is their I2V formulation: they can only access information from the first frame, preventing them from handling content that appears later in the video, camera viewpoint changes, or complex temporal reordering. Our V2V approach, in contrast, can leverage information from any frame of the input video.

\section{Ablations}

% \subsection{Jitter for Preventing Identity Copying}

\begin{figure}[h]
    \centering
    \includegraphics[width=1\linewidth]{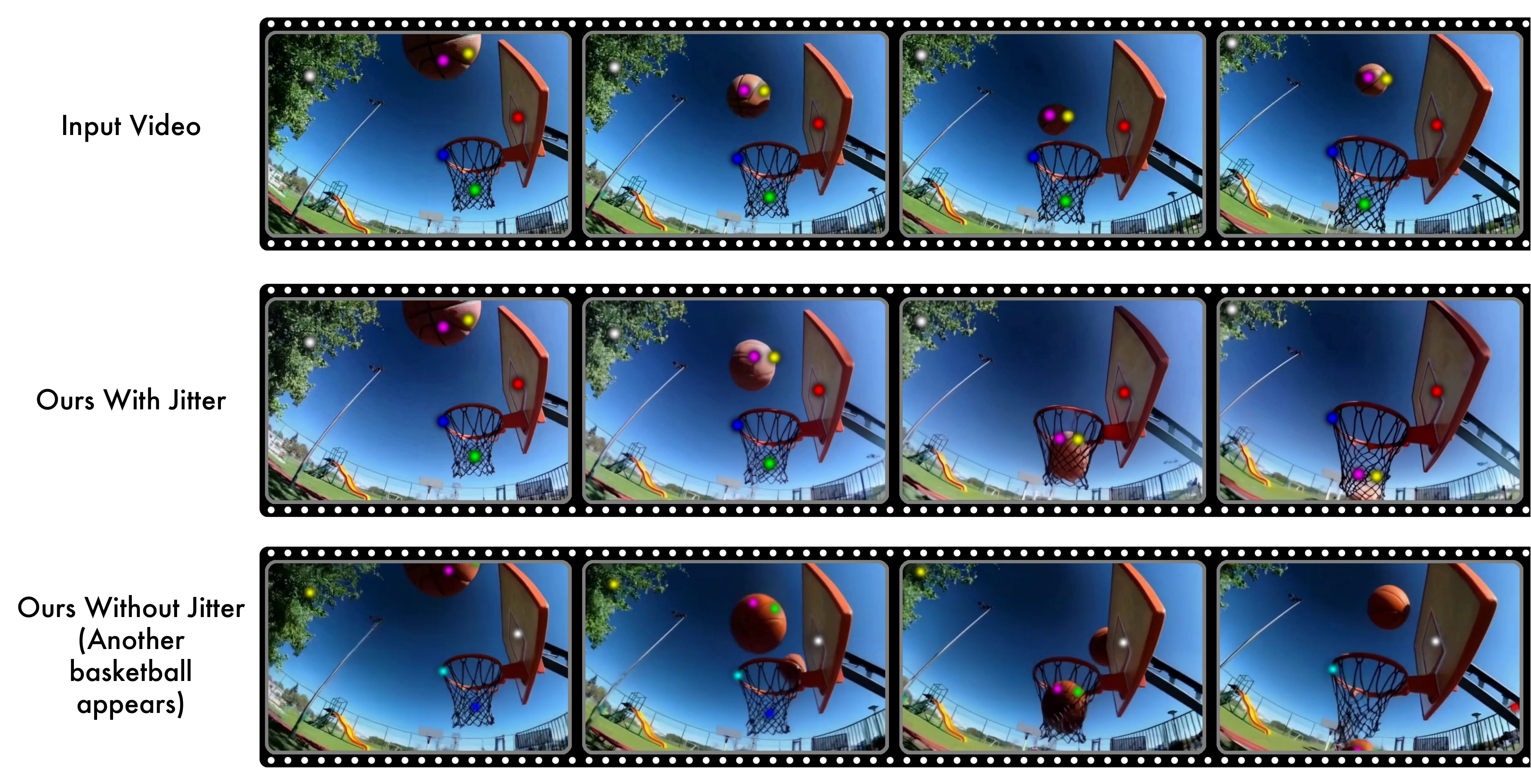}
    \caption{\textbf{The effects of trajectory jitter on motion editing.} Top: without jitter, a second basketball appears. Bottom: with 1-2 pixel jitter, the edit follows correctly.}
    \label{fig:jitter}
\end{figure}

We discovered an interesting phenomenon during inference: when tracking points are pixel-perfectly aligned with the input video trajectories across multiple frames, the model exhibits a strong bias toward reproducing the original video's semantics rather than following the edited motion.

Figure~\ref{fig:jitter} illustrates this effect. In the top row (without jitter), when tracking points are pixel-perfectly aligned with the input video, the model exhibits a bias toward reproducing the original video's semantics. Although the basketball successfully goes through the hoop following the edited trajectory, a second basketball mysteriously appears behind the hoop to match the original video where the basketball passes in front. This occurs because the pixel-perfect alignment of other tracking points signals to the model that it should preserve the content of the entire input video, which is often the only case where the points are aligned that perfectly during training. In the bottom row (with jitter), this identity-copying behavior is eliminated and the edit follows the intended motion correctly. 

To address this, we introduce a simple but effective inference-time technique: ``Jitter''. We add small random noise $\epsilon \sim \mathcal{U}(-2, 2)$ pixels to the $(x, y)$ positions of all tracking points at each frame. Importantly, this is an inference-time modification only—the model is not trained with this jitter. This minimal perturbation (1-2 pixels) is imperceptible in the rendered tracks but sufficient to break the model's tendency to copy the input video's identity, allowing it to follow the edited trajectories more faithfully.

\section{Future Work}

In future work, we consider creating large-scale synthetic datasets with precise motion counterfactuals made with 3d software. While our current approach leverages real videos paired with diffusion-generated conuterfactuals, synthetic 3d data would provide perfect ground truth motion-edit pairs, enabling exact control over individual object trajectories, physical interactions, and the resulting lighting and shading changes. This would improve the precision of our training dataset, possibly allowing even less points to be used for control.

% Commented out CVPR boilerplate sections
% \section{Rationale}
% \label{sec:rationale}
% %
% Having the supplementary compiled together with the main paper means that:
% %
% \begin{itemize}
% \item The supplementary can back-reference sections of the main paper, for example, we can refer to \cref{sec:intro};
% \item The main paper can forward reference sub-sections within the supplementary explicitly (e.g. referring to a particular experiment);
% \item When submitted to arXiv, the supplementary will already included at the end of the paper.
% \end{itemize}
% %
% To split the supplementary pages from the main paper, you can use \href{https://support.apple.com/en-ca/guide/preview/prvw11793/mac#:~:text=Delete%20a%20page%20from%20a,or%20choose%20Edit%20%3E%20Delete).}{Preview (on macOS)}, \href{https://www.adobe.com/acrobat/how-to/delete-pages-from-pdf.html#:~:text=Choose%20%E2%80%9CTools%E2%80%9D%20%3E%20%E2%80%9COrganize,or%20pages%20from%20the%20file.}{Adobe Acrobat} (on all OSs), as well as \href{https://superuser.com/questions/517986/is-it-possible-to-delete-some-pages-of-a-pdf-document}{command line tools}.

\end{document}